\documentclass{article}
\usepackage{spconf,amsmath,graphicx}

\usepackage{xcolor}

\usepackage{amssymb}
\usepackage{pifont}
\usepackage{hyperref}
\usepackage{amsmath}

\DeclareMathOperator*{\argmin}{arg\,min}

\usepackage{enumitem}
\setlist{nosep, leftmargin=14pt}

\usepackage{mwe} 

\title{Skeletal Point Representations with Geometric Deep Learning}
 \twoauthors
 {Ninad Khargonkar\sthanks{Work done while an intern at Kitware.\newline Code available at \url{https://github.com/kninad/skeleton-nn}}}
	{The University of Texas at Dallas\\
	Department of Computer Science\\
	Richardson, Texas}
 {Beatriz Paniagua, Jared Vicory}
	{Kitware, Inc\\
	Medical Computing\\
	Carrboro, North Carolina}
\begin{document}
%
\maketitle
\begin{abstract}
Skeletonization has been a popular shape analysis technique that models both the interior and exterior of an object. Existing template-based calculations of skeletal models from anatomical structures are a time-consuming manual process. Recently, learning-based methods have been used to extract skeletons from 3D shapes. In this work, we propose novel additional geometric terms for calculating skeletal structures of objects. The results are similar to traditional fitted s-reps but but are produced much more quickly. Evaluation on real clinical data shows that the learned model predicts accurate skeletal representations and shows the impact of proposed geometric losses along with using s-reps as weak supervision.
\end{abstract}


\section{Introduction}
\label{sec:intro}
\begin{figure}[!t]
	\centering
	\includegraphics[width=0.9\columnwidth]{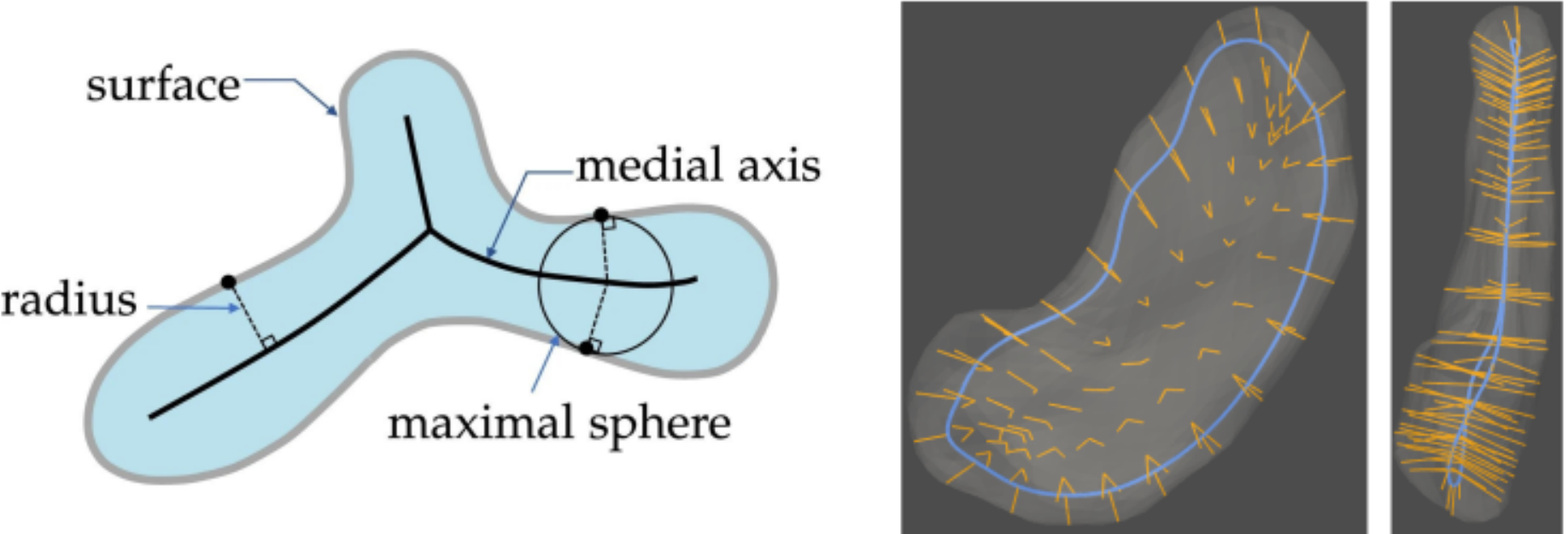}
	\caption{(Left) Medial axis for a 2D shape, (Right) s-rep for a hippocampus
	         surface with yellow lines as the spoke vectors.}
	\label{fig:teaser}
\end{figure}

Skeletonization has been a popular approach for modeling anatomical structures due to the ability of skeletal representations to capture both the boundary and interior of an object. This is in contrast with simpler models such as densely-sampled boundary landmarks. The most widely used approach to define an object's skeleton is through a medial axis transform~\cite{blum1967transformation-mat-org} (MAT)(fig. \ref{fig:teaser}, left). The MAT consists of a set of points and associated radii, called spokes, which form the set of maximally inscribed spheres inside the shape. MAT-based models have been used for a wide range of applications~\cite{tagliasacchi20163d-skel-survey} such as segmentation, registration, finite element modeling, and statistics of object shape.

The main limitation of the MAT is that it has the tendency to amplify small-scale noise on an object's boundary into large deviations in skeleton location and topology across a population. This makes it unsuited for uses which require consistent structure such as statistical analysis.

This limitation has led to modifications~\cite{yan2018voxel-matapprox, skel3d-wu2015-DPC, skel3d-huang2013l1-medial} to the MAT to address these shortcomings. In particular, \textit{s-reps}~\cite{pizer2020-srep-object}(fig. \ref{fig:teaser}, right) are a class of discrete skeletal representations related to the MAT which have a fixed topology and consistent sampling across a population. This is done by fitting a template s-rep to an object via an optimization~\cite{liu2021-srep-fitting} rather than direct computation from the object's boundary. Having a fixed template which is then adapted to each individual in a population yields improved consistency and correspondence. 

The optimization is constrained in such a way as to have the final object be nearly medial in that points on the skeleton are approximately equidistant from the top and bottom of the object's surface and the radii emanating from these points are nearly orthogonal to the boundary. This allows for s-reps to have the stability required for statistical analysis while still taking advantage of some of the beneficial properties of a purely medial skeleton. A downside to this approach is that the optimization can be slow and often requires manual template generation and parameter tuning when applied to a new dataset.

Learning-based methods for skeletonizing images and shapes are a relatively recent line of 
research which have shown promise in robustly computing skeletal representations. Earlier
learning-based methods primarily focused on the task of extracting skeletons from 2D
natural images~\cite{skel2d-shen2017deepskeleton}. They typically
involved use of fully convolutional architectures and pose the problem in a similar
fashion to that of image segmentation as seen in works based on the popular U-Net 
architecture~\cite{skel2d-panichev2019u, skel2d-nguyen2021u}. In contrast, there is less work on learning-based
skeletonization of 3D objects, partially due to their increased complexity and partially due 
to the lack of a benchmark dataset for training. This led to the development of point-based 
methods like Point2Skeleton~\cite{skel3d-p2skel-lin2021} which utilizes PointNet++~\cite{qi2017pointnet++}
as a point encoder and tries to predict weights on input points to generate the skeleton as a convex combination 
of the inputs in a manner similar to~\cite{skel3d-chen2020unsupervised}.

\begin{figure*}[!t]
	\centering
	\includegraphics[width=1.6\columnwidth]{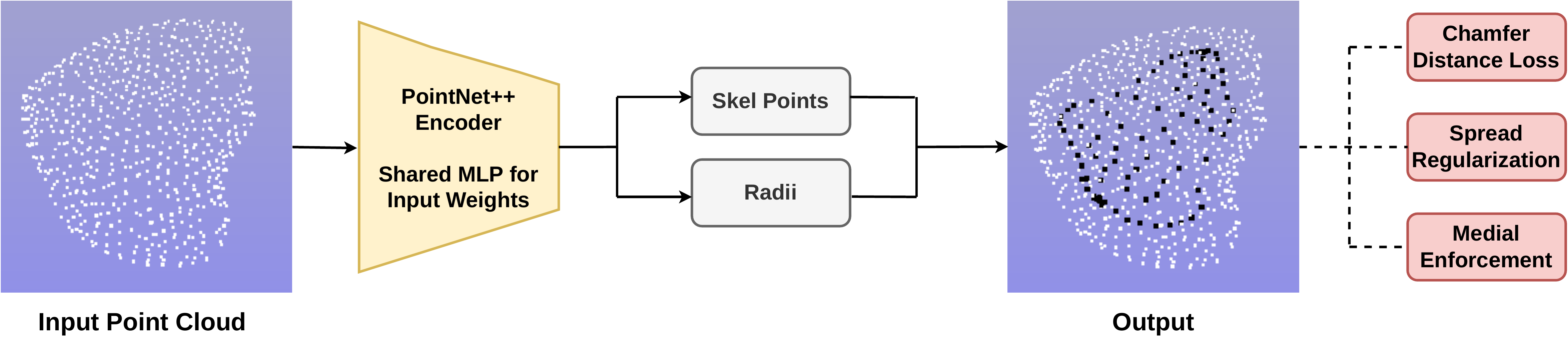}
	\caption{Model overview and geometric terms used for training.}
	\label{fig:model}
\end{figure*}

One of the drawbacks of the point based methods is that they are not optimal for use in shape
analysis tasks like s-rep fitting due to inconsistent sampling density and sometimes large differences in the MAT of similar shapes.
Such deviations are harder to tolerate for applications in shape analysis where a consistent pattern in the skeletons across a population is necessary. 
In this paper we build upon these point-based learning methods by introducing a novel way to utilize s-reps as
a form of supervision (instead of computing MATs) and introducing two
additional geometric loss terms, one for enforcing the medial approximation of the skeletal
points and a second regularization term to encourage spread between them. The proposed changes
show good results on synthetic data as well as promising application to two clinical problems of modeling hippocampi and tricuspid valve leaflets.

\section{Methodology}
\label{sec:method}

In this section we present the base model that defines a learning based s-rep to then motivate the
three geometric terms for better skeletal point prediction.

\subsection{Base Model}
\begin{figure}
	\centering
	\includegraphics[width=0.8\columnwidth]{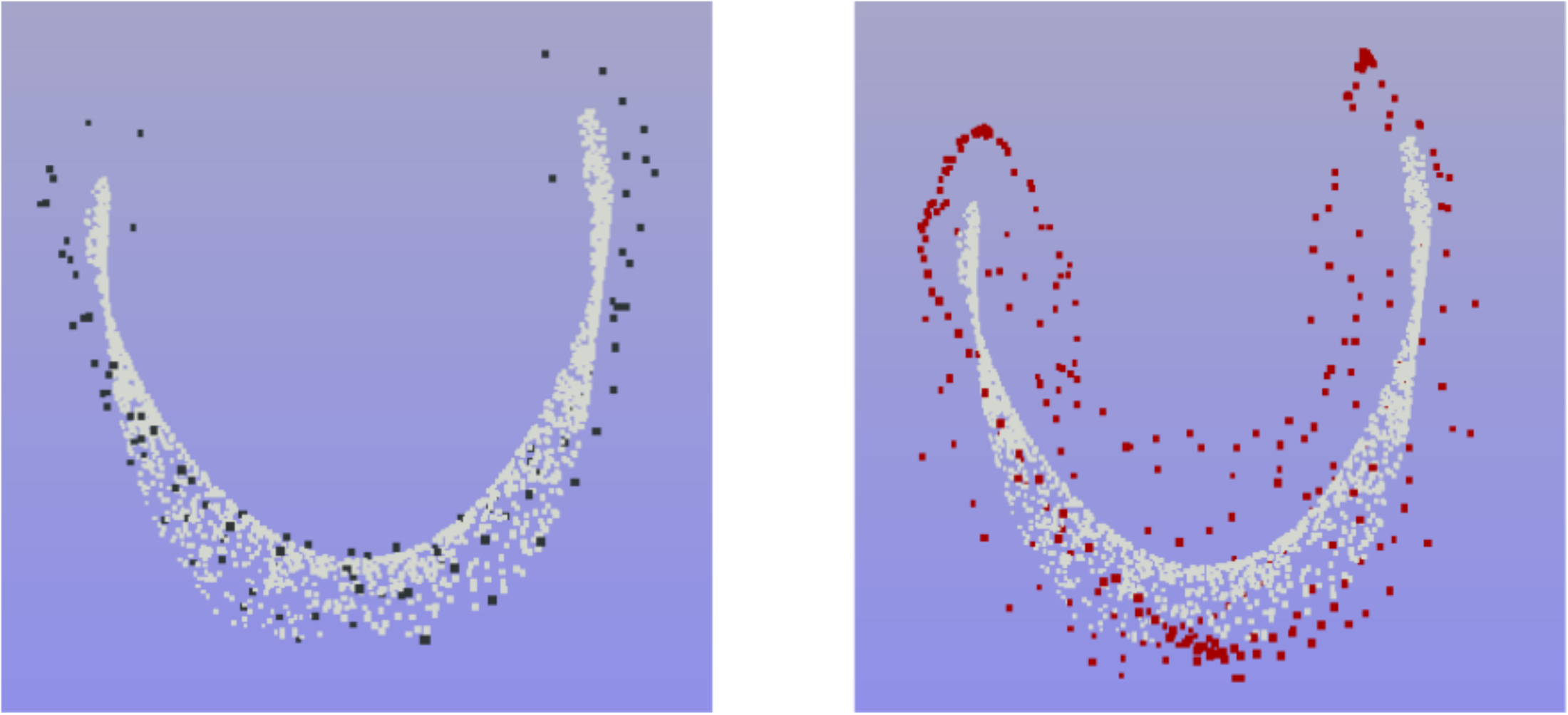}
	\caption{Effect of s-rep supervision: White dots are the G.T skeletal points, black are the
	predictions with supervision and red are without supervision.}
	\label{fig:unsup-vs-sup}
\end{figure}

Our basic definition of a learning-based s-rep is built upon the previous Point2Skeleton~\cite{skel3d-p2skel-lin2021, skel3d-chen2020unsupervised} method.
As seen in fig.~\ref{fig:model}, the input for our model is a segmented shape's surface point cloud. We then
use the PointNet++ layers as encoders and have a shared multilayer perceptron (MLP) across all input points to predict weights over them. Assuming there are N input points $P$ after the encoding and M skeletal points $S$ to be
predicted, the model will try to predict a weight matrix $W \in \mathcal{R}^{N \times M}$. The weights
are constrained to form a convex combination over the input points and hence we obtain the skeletal
points $S = W^T P$. The radii are predicted in a similar fashion by first computing
the distance $d_i = min_{s_j \in S} ||p_i - s_j||_2 , \;\forall p_i \in P$. Then the radius for each
skeletal point is predicted using $r(s_j) = \sum_{i=1}^{N} W_{ij} d_i$ using the convex combination over
the input points. The drawback with this approach is that the model does not learn to well approximate a smooth skeletal sheet due to the lack of supervision. The predictions are  scattered around the interior as seen on the right of fig.~\ref{fig:unsup-vs-sup}.

\subsection{Novel Geometric Losses}
\subsubsection{MAT supervision via s-reps}
We propose to use supervision to allow the network to find skeletal points which better approximate a smooth sheet. Using the MAT for supervision will yield a network susceptible to producing skeletons with the same undesirable properties. We instead propose to use pre-computed s-reps as a form of supervision to allow the model to better approximate actual s-reps. 
 The supervision with s-reps as ground truth is enforced through the means of a symmetric Chamfer distance loss ($L_{CD}$) with the predicted skeletal points as seen in Eqn.~\ref{eqn:sup-chamfer-dist}, where $S1$ and $S2$ are the predicted and ground truth skeletal sheets. 
\begin{equation}
\begin{split}
\label{eqn:sup-chamfer-dist}
    L_{CD}(S_1, S_2) = \frac{1}{|S_1|} \sum_{a \in S_1} \min_{b \in S_2} ||a - b||_2^{2} \\
                     + \frac{1}{|S_2|} \sum_{b \in S_2} \min_{a \in S_1} ||b - a||_2^{2}
\end{split}
\end{equation}

\subsubsection{Spread regularization}
Another aspect of the base model is that there is no regularization on the skeletal points in terms
of their spread. Ideally we want the skeletal points to cover the interior of the object as well as 
possible. Without this, some parts of the object interior are not well represented which in turn
affects the quality of the surface reconstructed from the skeleton. Thus we impose a spread regularization
term $L_{spread}$ shown in Eqn.~\ref{eqn:spread} to discourage the clustering of skeletal points by penalizing
the average pairwise distances between them.
\begin{equation}
\label{eqn:spread}
 L_{spread} = -\frac{1}{M^2} \sum_{i=1}^{N}\sum_{j=1}^{M} ||s_i - s_j||_2
\end{equation}

\subsubsection{Medial Enforcement}
Recall that the skeletal points in a MAT have more than one closest point on the 
surface. The base model incorporates this loss through a point to sphere 
loss~\cite{skel3d-p2skel-lin2021} with two terms: one on the input points forcing them to lie on the closest skeletal surface and another on the skeletal points forcing them to touch the closest
input point. We modify the second to encourage a medial approximation by having each skeletal
point's sphere touch its nearest three points following the definition of MAT. This
loss $L_{medial}$, shown in Eqn.~\ref{eqn:medial} will help train the network to adhere to a stronger 
medial approximation.
\begin{equation}
\label{eqn:medial}
  L_{medial} = \sum_{s \in S} \sum_{p \in C_3(s)} || p - s||_2 - r(s)
\end{equation}
Here the set $C_3(s) \subset P$ represents the closest three input surface points to the skeletal point 
$s$ and $r(s)$ represents the radius for the skeletal sphere at $s$.

\subsubsection{Spoke vectors}
To ensure a fair comparison with s-reps and obtain a surface reconstruction we also leveraged a simple
way to predict the spoke vectors toward the surface for each skeletal point from the weight 
matrix.
Intuitively the surface points closer to the skeletal points have a higher weight and hence hold clues to
which direction the nearest surface location lies for a skeletal point. Therefore the spoke vectors
are predicted to be the direction vectors to the highest weight input point (for each skeletal point)
as shown in Eqn.~\ref{eqn:spokes}
\begin{equation}
\label{eqn:spokes}
    spoke(s_j) = \frac{(p_{i^*_j} - s_j)}{||p_{i^*_j} - s_j||_2} \; \text{where } i^*_j = \argmin_{i=1,\dots,N} W_{ij} 
\end{equation}
Note that all surface reconstruction results utilize the notion of spoke vectors which is a bit different
from earlier works where either (a) uniform sampling is used on the skeletal sphere, or (b) the merged
mesh from all skeletal spheres is used to reconstruct the surface.

\section{Experiments}
\label{sec:experiments}

\subsection{Implementation}
We used PyTorch version 1.12 for our model implementation running on
CUDA 11.6. The experiments were performed on a machine with a single NVIDIA RTX A5000 GPU having 
24GB vRAM. The Adam optimizer was used for training with a learning rate
of 5e-4 and a batch size of 6. We followed a similar training regime as that of 
\cite{skel3d-p2skel-lin2021} for enabling the geometric losses on the model. For both training and inference, 
1000 points were randomly sampled from the normalized input 
(zero mean, scaled down to unit sphere). We set 100 
skeletal points as the desired output from the model. The s-reps for the hippocampi and leaflets
data were obtained using the skeletal representation module in SlicerSALT~\cite{vicory2018slicersalt}, a toolbox of shape analysis software.
For the quantitative evaluation metrics, we utilized the symmetric versions of the Chamfer and
Hausdorff distances between two point sets abbreviated as
C.D and H.D respectively in Table~\ref{table:ablation-study} and Table~\ref{table:evaluation}.

In our implementation, predicting an s-rep takes approximately 1.7 seconds. The time needed for the s-rep optimization in SlicerSALT varies depending on how much refinement of the initial s-rep fit is needed. Typical runs take between 5 and 30 minutes but runs over an hour occasionally occur.

\subsection{Synthetic Dataset}
\label{ssec:synthetic}
\begin{figure}[!t]
	\centering
	{\includegraphics[width=0.8\columnwidth]{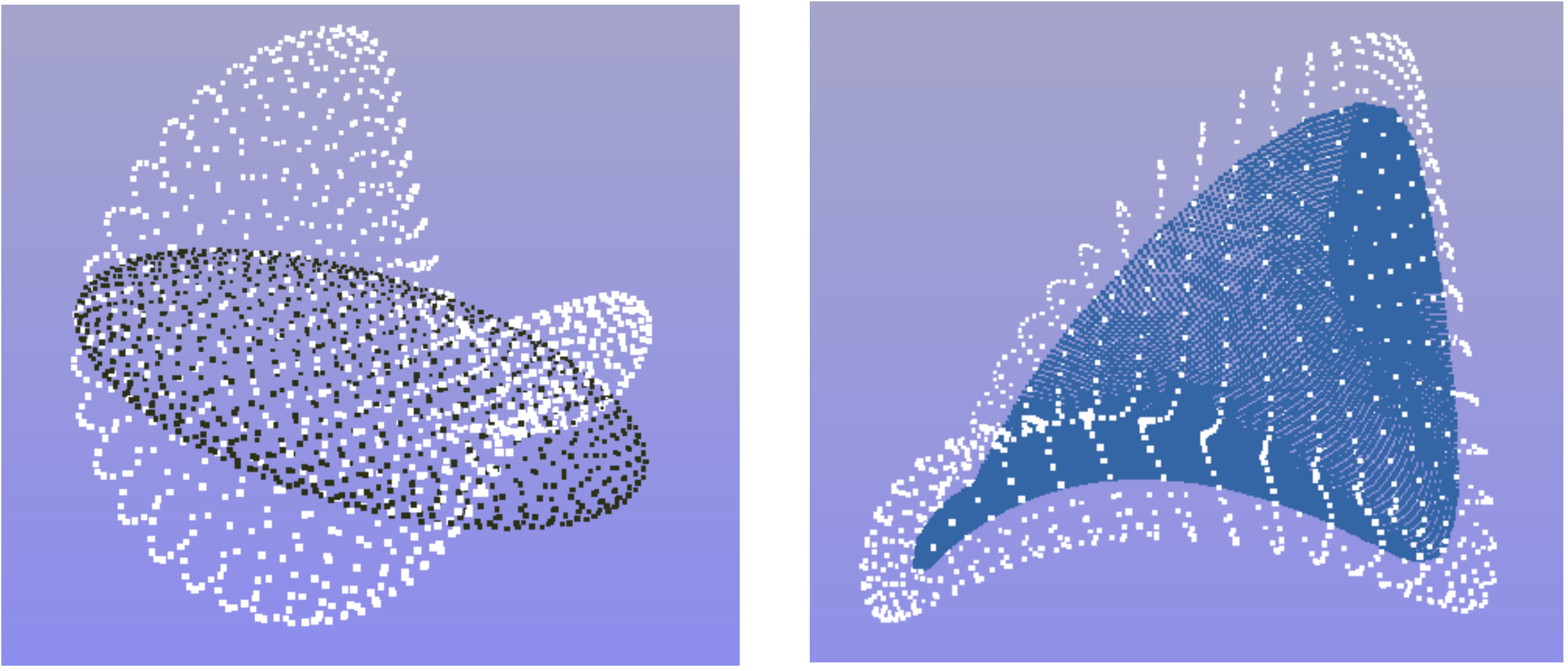}}
	{\includegraphics[width=0.8\columnwidth]{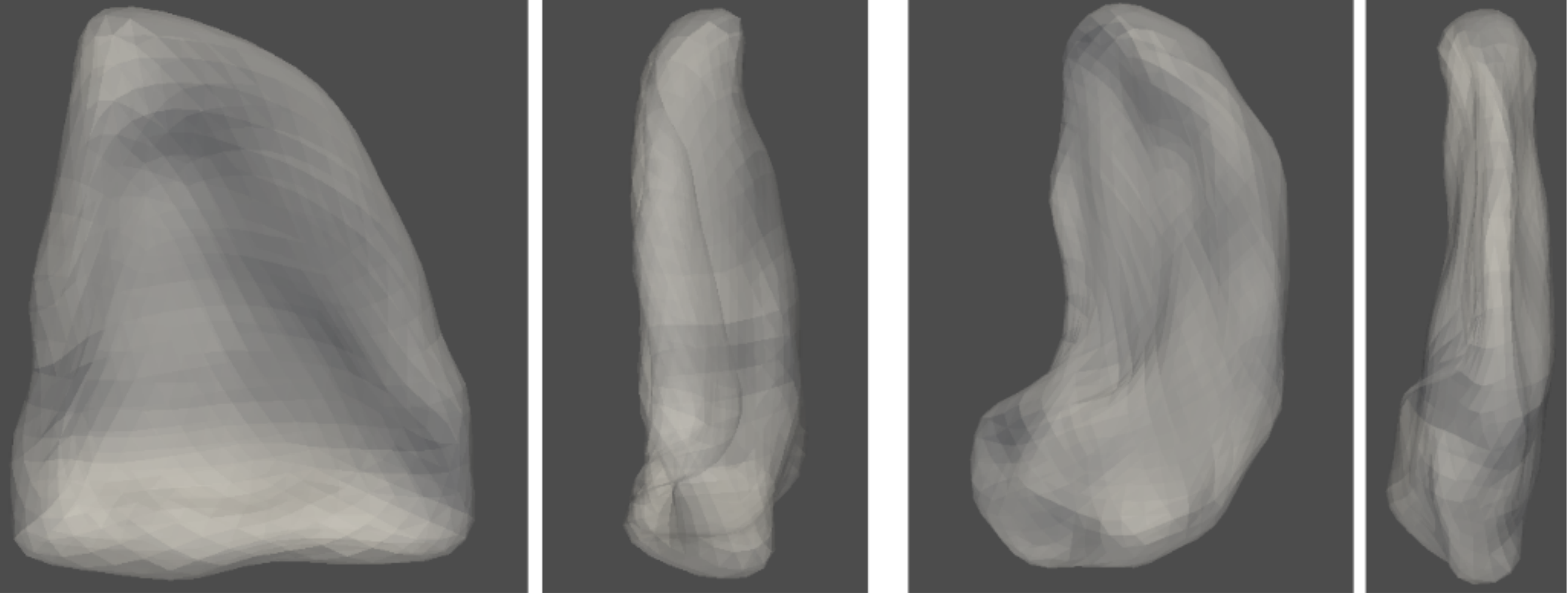}}
	\caption{ (Top) The left image shows an ellipsoid (black) and a version deformed into a leaflet (white) while the right image shows the ground truth skeletal points (blue sheet). (Bottom) Sample images (front/side view) of leaflets and hippocampi.}
	\label{fig:data-imgs}
\end{figure}

As stated before, there is no standard 3D dataset for the task of skeletal point prediction.
Taking inspiration from current s-rep fitting approaches which rely on deforming ellipsoids to match an object, we generated a synthetic dataset of deformed ellipsoids along with their s-reps. Starting from a base ellipsoid and s-rep which are axis aligned, we first apply random scale factors sampled from the normal distribution $\mathcal{N}(1,0.15)$ to each axis independently. We then deform the ellipsoid by bending the long axis by angles sampled from $\mathcal{N}(\frac{\pi}{3},\frac{\pi}{8})$ and twisting by angles sampled from $\mathcal{N}(\frac{\pi}{6},\frac{\pi}{8})$. Fig.~\ref{fig:data-imgs} shows an example of an ellipsoid pre and post deformation. 

\subsection{Ablation Study}
We conducted an ablation study of the proposed geometric terms on 5000 deformed ellipsoids generated as described in \ref{ssec:synthetic}. Table~\ref{table:ablation-study} shows the performance of the individual and combined
effects on held out data. We refer to using the base model with the supervised loss as the
baseline method. The ablation experiments evaluated two aspects of the model, the
first measuring how close the predicted skeletal points match the s-rep skeletal points and the
second measuring the effectiveness of surface reconstruction using the spoke vectors.
\begin{table}[h!]
\centering
\caption{Ablation study results on different configurations.}
\label{table:ablation-study}
\begin{tabular}{cc|cc|cc}
\hline \hline
\multicolumn{2}{l|}{Loss Terms} & \multicolumn{2}{l|}{G.T. Skel Points} & \multicolumn{2}{l}{Surf Recon} \\ \hline
Spread & Medial & C.D    & H.D    & C.D    & H.D    \\
\hline
\ding{55} & \ding{55} & 0.0085 & 0.3528 & 0.0395 & 0.4122 \\
\ding{51} & \ding{55} & 0.0089 & 0.2082 & 0.0297 & 0.4157 \\
\ding{55} & \ding{51} & 0.0097 & 0.3527 & 0.0385 & 0.4396 \\
\ding{51} & \ding{51} & \textbf{0.0073} & \textbf{0.1903} & \textbf{0.0247} & 0.4366 \\
\hline
\end{tabular}
\end{table}

As we can see in table \ref{table:ablation-study}, the combined effect of both the terms tends to give the best result. The effect of the spread regularization is more noticeable than that of the 
medial term, perhaps due to the fact that the base model also enforces the medial terms albeit less
strongly. The surface reconstruction results with the spoke vectors under the H.D metric are weaker than those of the skeleton due to limitations with predicted spoke vector orientation.

\subsection{Clinical Data Evaluation}
We evaluated the model on shape data from hippocampi and heart valve leaflets with boundaries computed via SPHARM-PDM.
Since the real world shape data for such anatomical structures is quite limited, we
utilized a fine-tuning strategy. First we trained the model using the synthetic deformed
ellipsoid data to enable learning on a wide variety of shapes. Using this pre-trained model,
we fine-tuned using the limited amount of hippocampi or leaflet data.
Similar to the ablation study, we compared the skeletal point predictions with the s-rep skeleton points 
along with the surface reconstruction quality. Additionally, we also evaluated how similar is the surface
reconstruction to that derived from the s-rep. 
\begin{table}[h!]
\centering
\caption{Results on Hippocampi and Leaflet data}
\label{table:evaluation}
\begin{tabular}{c|cc|cc|cc}
\hline \hline
\centering
Data & \multicolumn{2}{c|}{Skel. Points} & \multicolumn{2}{c|}{Surf Recon} & \multicolumn{2}{c}{Srep Recon} \\ \hline
           & C.D    & H.D    & C.D    & H.D    & C.D    & H.D    \\
           \hline
Hipp. & 0.004 & 0.097 & 0.014 & 0.238 & 0.023 & 0.231 \\
Leaf.   & 0.007 & 0.146 & 0.025 & 0.097 & 0.044 & 0.275 \\
\hline    
\end{tabular}
\end{table}

\begin{figure}[!h]
	\centering
	\includegraphics[width=0.8\columnwidth]{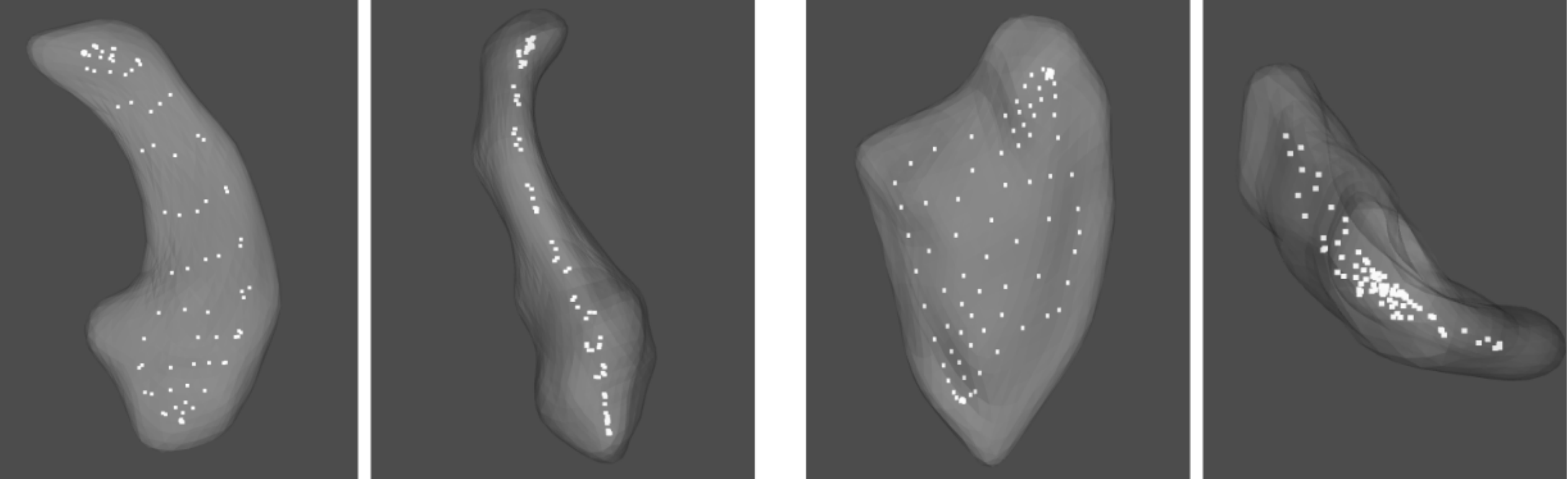}
	\caption{Qualitative results on the hippocampi (left) and leaflets (right) showing front and side views for both. The white dots are the predicted skeletal points.}
	\label{fig:qual-results}
\end{figure}

Table.~\ref{table:evaluation} summarizes the results on 
the held out samples for each dataset. As we can see, the predicted skeletons are close
to what we would expect after the s-rep fitting process while also being able to reconstruct
the surface fairly well even though the spoke vectors are not refined. Fig.~\ref{fig:qual-results}
shows qualitative results of the skeletal point prediction on the two datasets. The figure
shows well spread out skeletal points along with them lying a sheet-like surface.

\subsection{Discussion}
\label{subsec:discussion}
The experiments shows promising results in obtaining skeletal representations for slabular structures. Due to its learning-based nature, its runtime has an advantage over the s-reps fitting process.
It is important to note that there is a trade-off since s-reps may be more accurate for
surface reconstruction due to principled computation of spoke vectors. This shows one of the
limitations of point based models since we do not account for the surface normal information,
which could help align the spoke vectors to the surface. Adapting the model to work with surface
meshes may also refine the results after taking into account other geometric features like curvature.

A limitation of the point based models is the requirement to specify the number of output skeletal points while training. This could result in the the skeletal surface having a lesser density of predicted points. Spline-based interpolation methods could help here to obtain a dense and continuous 
representation for the  skeleton. Finally, in the context of real world shape analysis applications, it 
is important to note that all skeletonization methods assume a segmented boundary surface. For 
anatomical structures, it is likely that the surface is the result of an algorithm that will
be uncertain on certain parts of the surface. Therefore, future work could involve incorporating the
uncertainty into the skeletal point prediction, specifically while assigning weights to input points.   


\section{Conclusion}
\label{sec:conclusion}
In this paper we presented novel technique to leverage recent point-based
learning methods for 3D shape skeletonization. We showed the relative
benefits of utilizing s-reps for supervision and the impact of two
new geometric losses for training more appropriate skeletons. The results on
real world data show promise in terms of both qualitative and quantitative
measures and also point towards future research in terms of adapting
to input modality and surface reconstruction.


\section{Compliance with Ethical Standards}
\label{sec:compliance-ethical}
Heart leaflets from the tricuspid valve were obtained from transthoracic 3DE images. This modality is part of a standard clinical echo lab protocol for children with congenital heart disease at Children's Hospital of Philadelphia. 

The hippocampus data was provided by Martin Styner, UNC Neuro Image Analysis Laboratory,
and is publicly available as part of the SlicerSALT and SPHARM-PDM tool distributions. Original data acquisition was funded by the Stanley Foundation.

Both studies were performed according to a protocol approved by the
institutional review board at the relevant institutions.

\section{Acknowledgments}
\label{sec:acknowledgments}

This research was supported by the NIH under project R01EB021391. We thank Matthew A. Jolley of Children's Hospital of Philadelphia for lending the clinical heart valve leaflet data for validation of the algorithms presented here.  Ninad Khargonkar is currently affiliated to the University of Texas at Dallas but the work contained in this manuscript was performed while he was an intern at Kitware. The authors have no other financial or non-financial interests to disclose.

\bibliographystyle{IEEEbib}
\bibliography{refs}

\end{document}